\documentclass[letterpaper, 10 pt, conference]{ieeeconf}
\IEEEoverridecommandlockouts
\usepackage{cite}
\usepackage{amsmath,amssymb,amsfonts}
\usepackage{algorithmic}
\usepackage{algorithm}
\usepackage{graphicx}
\usepackage{textcomp}
\usepackage[capitalize]{cleveref}
\usepackage{booktabs}
\usepackage{xcolor}
\def\BibTeX{{\rm B\kern-.05em{\sc i\kern-.025em b}\kern-.08em
    T\kern-.1667em\lower.7ex\hbox{E}\kern-.125emX}}
\begin{document}

\title{\LARGE \bf
GoalVLM: VLM-driven Object Goal Navigation for Multi-Agent System
}

\author{MoniJesu James, Amir Atef Habel, Aleksey Fedoseev, and Dzmitry Tsetserokou
\thanks{The authors are with the Intelligent Space Robotics Laboratory, Center for Digital Engineering, Skolkovo Institute of Science and Technology, Moscow, Russia. 
\tt \{monijesu.james, amir.habel, aleksey.fedoseev, d.tsetserokou\}@skoltech.ru}
}


\maketitle

\begin{abstract}

Object-goal navigation has traditionally been limited to ground robots with closed-set object vocabularies. Existing multi-agent approaches depend on precomputed probabilistic graphs tied to fixed category sets, precluding generalization to novel goals at test time.

We present GoalVLM, a cooperative multi-agent framework for zero-shot, open-vocabulary object navigation. GoalVLM integrates a Vision-Language Model (VLM) directly into the decision loop, SAM3 for text-prompted detection and segmentation, and SpaceOM for spatial reasoning, enabling agents to interpret free-form language goals and score frontiers via zero-shot semantic priors without retraining. Each agent builds a BEV semantic map from depth-projected voxel splatting, while a Goal Projector back-projects detections through calibrated depth into the map for reliable goal localization. A constraint-guided reasoning layer evaluates frontiers through a structured prompt chain (scene captioning, room-type classification, perception gating, multi-frontier ranking), injecting commonsense priors into exploration.

We evaluate GoalVLM on GOAT-Bench val\_unseen (360 multi-subtask episodes, 1032 sequential object-goal subtasks, HM3D scenes), where each episode requires navigating to a chain of 5--7 open-vocabulary targets. GoalVLM with $N=2$ agents achieves 55.8\% subtask SR and 18.3\% SPL, competitive with state-of-the-art methods while requiring no task-specific training. Ablation studies confirm the contributions of VLM-guided frontier reasoning and depth-projected goal localization.

\end{abstract}
\vspace{3pt}

\section{Introduction}
Autonomous robot navigation has evolved from simple point-to-point geometric pathfinding to high-level semantic goal achievement. Object-goal navigation represents a significant challenge, requiring an agent to autonomously explore an unfamiliar environment to locate an instance of a specified object category~\cite{yuan2024gamapzeroshotobjectgoal}. Recent benchmarks such as GOAT-Bench~\cite{khanna2024goatbenchbenchmarkmultimodallifelong} extend this to sequential multi-subtask episodes with open-vocabulary goals, while multi-agent approaches offer faster coverage through cooperative exploration~\cite{huang2025expanding}. However, most prior work is limited to single robots or fixed instruction sequences with closed-set categories.
We propose GoalVLM to address this gap: a cooperative multi-agent system for open-vocabulary ObjectNav. By combining advanced vision-language perception with distributed map sharing, GoalVLM enables zero-shot goal understanding and efficient exploration. Our contributions include:
(1)~A VLM-driven multi-agent coordination strategy that dynamically allocates exploration frontiers, with agents sharing fused semantic maps for global scene awareness;
(2)~A zero-shot perception pipeline using SAM3 for text-prompted detection and a GoalProjector that back-projects detections through calibrated depth into the BEV map for reliable goal localization;
(3)~A constraint-guided reasoning layer where SpaceOM evaluates candidate frontiers through structured prompt chains incorporating scene captioning, room-type classification, and multi-frontier ranking;
(4)~Evaluation on GOAT-Bench with ablation studies quantifying the contribution of VLM reasoning and goal projection.

\begin{figure}[t]
\centering
\includegraphics[width=1.0\linewidth]{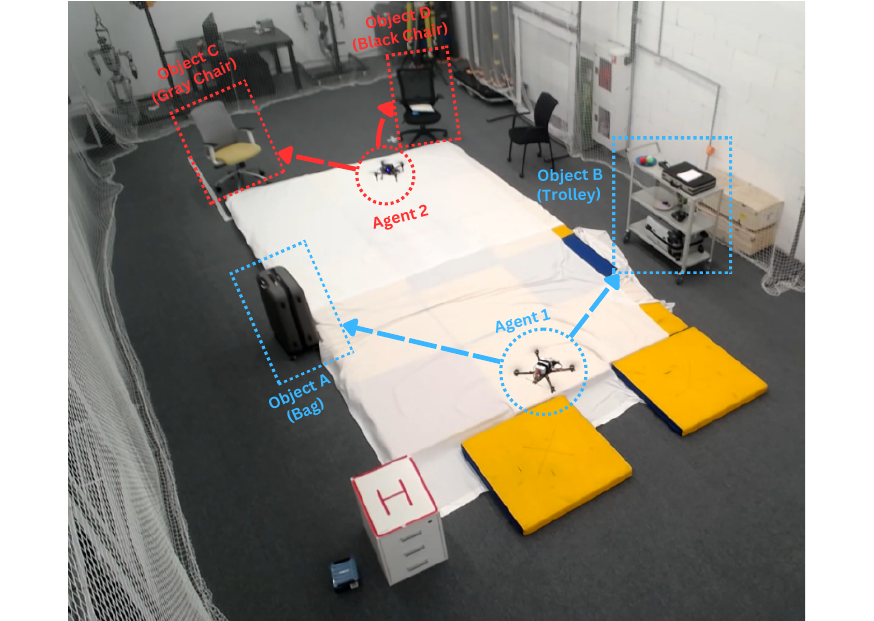} 
\caption{Real Experiment of Multi-Agents of GoalVLM.}
\label{fig:title}
\vspace{-0.5cm}
\end{figure}

\section{Related Work}
\subsection{Multi-Agent Navigation}
Coordinated exploration and navigation have been studied in multi-robot systems. Classical frontier-based exploration identifies boundary frontiers between known free space and unknown areas to plan exploration targets~\cite{613851}. Multi-robot extensions enable independent agents to divide labor by sharing maps and frontiers. Recent work shows that sharing semantic maps via distributed relational networks yields near-linear speedup in object search. MCoCoNav~\cite{yu2023co} introduces multi-agent cooperative navigation with LLM-based coordination, while Co-NavGPT~\cite{yu2023co} and MCOCONAV \cite{shen2025enhancingmultirobotsemanticnavigation} leverage GPT and GLM4V, respectively, for frontier reasoning. Unlike past methods limited to homogeneous ground robots, GoalVLM combines multiple agents with VLM-driven spatial reasoning to efficiently explore and locate open-vocabulary targets.

\begin{figure*}[t!]
    \centering
    \includegraphics[width=0.8\linewidth]{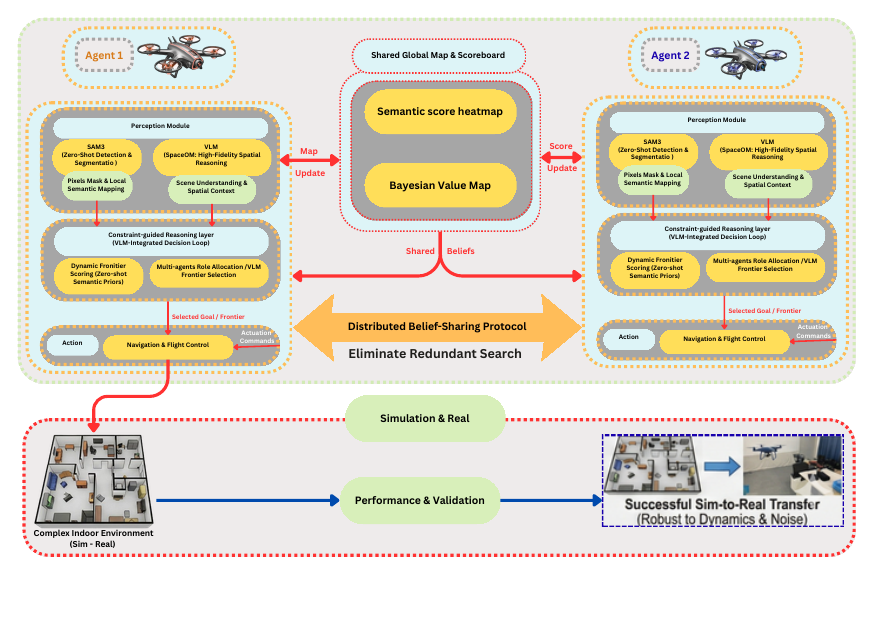}
    \caption{Multiple agents process vision-language cues, perform local planning, and share semantic maps.}
    \label{fig:architecture}
\end{figure*}

\subsection{Vision-Language Models for Robotics}
Vision-Language Models (VLM) have enabled open-vocabulary perception in navigation. Models such as SAM~\cite{kirillov2023segment} and Grounding DINO~\cite{ren2024grounding} provide zero-shot object detection and segmentation, allowing agents to respond to textual goals outside fixed training categories. VLFM~\cite{yokoyama2024vlfm} uses pretrained VLMs to compute a language-grounded value map over frontiers, and L3MVN~\cite{yu2023l3mvn} employs an LLM to choose frontier goals for exploration. GoalVLM similarly leverages text-prompted detection (SAM3) for visual grounding and a spatial reasoning model (SpaceOM) to encode commonsense priors (e.g., ``fridge'' $\rightarrow$ ``kitchen''). Unlike VLN methods that follow step-by-step instructions, our agents interpret high-level object labels and free-form descriptions, enabling direct ObjectNav without detailed instructions.

\subsection{Open-Vocabulary Object Navigation}
ObjectNav traditionally assumes a fixed set of object classes. The HM3D-OVON benchmark~\cite{yokoyama2024hm3d} broadens this to open-vocabulary goals (379 categories via free-form language). GOAT-Bench~\cite{khanna2024goatbenchbenchmarkmultimodallifelong} further extends this to sequential multi-subtask episodes where an agent must navigate to a chain of open-vocabulary targets. ApexNav~\cite{zhang2025apexnav} and VLFM~\cite{yokoyama2024vlfm} illustrate how semantic cues can direct exploration. GoalVLM extends these ideas to a multi-agent context: each agent performs zero-shot detection with SAM3 and queries SpaceOM to estimate $P(target | region)$ when selecting where to go, integrating both learned and commonsense priors.

\section{Method}

GoalVLM introduces a fully decentralized multi-agent framework for open-vocabulary object-goal navigation. As illustrated in Fig.~\ref{fig:architecture}, each agent executes a continuous closed-loop policy comprising perception, spatial reasoning, map fusion, and local planning. Agents share a fused BEV semantic map and coordinate frontier assignments to eliminate redundant exploration.

\subsection{Problem Formulation}
We formalize multi-agent object-goal navigation as a decentralized partially observable Markov decision process (Dec-POMDP). Given a set of $N$ agents $\mathcal{A} = \{a_1, \ldots, a_N\}$ operating in an unknown environment $\mathcal{E}$, the task is to locate an object instance belonging to a natural language specified category $g \in \mathcal{G}$, where $\mathcal{G}$ is an open vocabulary of 379+ categories.

At each timestep $t$, agent $a_i$ receives an observation $o_t^i = (I_t^{rgb}, I_t^{depth}, p_t^i)$ comprising an RGB image $I_t^{rgb} \in \mathbb{R}^{H \times W \times 3}$, a depth image $I_t^{depth} \in \mathbb{R}^{H \times W}$, and the agent's pose $p_t^i = (x, y, \theta)$ in the global coordinate frame. Each agent maintains a local semantic map $\mathcal{M}_t^i$ and executes actions from a discrete action space $\mathcal{A} = \{\texttt{STOP}, \texttt{MOVE\_FORWARD}, \texttt{TURN\_LEFT}, \texttt{TURN\_RIGHT}, \\ \texttt{TURN\_LEFT\_S}, \texttt{TURN\_RIGHT\_S}\}$. In the GOAT-Bench protocol, each episode contains a sequence of object-goal subtasks; an episode terminates successfully when the agent calls \texttt{STOP} within Euclidean distance $d_{success} = 1.0$\,m of each target object instance.

\subsection{Ego-Centric Semantic Mapping}
The semantic mapping module constructs a bird's-eye view (BEV) occupancy grid from ego-centric RGB-D observations. This involves three key transformations: camera projection, geometric transformation, and voxel splatting.

\subsubsection{Camera Intrinsic Model}
Given a depth image $I^{depth}$ captured at resolution $(W_s, H_s)$ with the horizontal field-of-view $\theta_{hfov}$, we compute the camera intrinsic matrix accounting for non-uniform rescaling to processing resolution $(W_p, H_p)$:
\begin{equation}
f_s = \frac{W_s}{2 \tan(\theta_{hfov}/2)}, \quad s_x = \frac{W_p}{W_s}, \quad s_y = \frac{H_p}{H_s}
\end{equation}
\begin{equation}
f_x = f_s \cdot s_x, \quad f_z = f_s \cdot s_y
\end{equation}
where $f_s$ is the sensor focal length (assuming square pixels), and $f_x, f_z$ are the rescaled horizontal and vertical focal lengths. This formulation correctly handles portrait-orientation sensors (e.g., $360 \times 640 \rightarrow 160 \times 120$) where $s_x \neq s_y$.

\subsubsection{Point Cloud Generation}
Each depth pixel $(u, v)$ with value $d_{uv}$ is back-projected to 3D camera coordinates:
\begin{equation}
\mathbf{P}_{cam} = \begin{bmatrix} (u - c_x) \cdot d_{uv} / f_x \\ d_{uv} \\ (v - c_y) \cdot d_{uv} / f_z \end{bmatrix}
\end{equation}
where $(c_x, c_y) = ((W_p - 1)/2, (H_p - 1)/2)$ is the principal point. The point cloud is then transformed to geocentric coordinates via a camera elevation rotation $\mathbf{R}_\phi$ and translation by sensor height $h_s$:
\begin{equation}
\mathbf{P}_{geo} = \mathbf{R}_\phi \mathbf{P}_{cam} + [0, 0, h_s]^T
\end{equation}

\subsubsection{Voxel Splatting and Height Slicing}
Points are accumulated into a voxel grid $\mathcal{V} \in \mathbb{R}^{V \times V \times Z \times C}$ via differentiable splatting, where $V$ is the vision range, $Z$ is the height discretization, and $C$ is the number of semantic channels. The obstacle map is obtained by summing voxels in the agent traversibility height band $[z_{min}, z_{max}]$:
\begin{equation}
\mathcal{M}_{obs}(x, y) = \sum_{z=z_{min}}^{z_{max}} \mathcal{V}(x, y, z, 0) > \tau_{obs}
\end{equation}
where $z_{min} = 25cm$ (above floor noise) and $z_{max} = h_s + 50cm$ (agent collision envelope). The explored area is computed by summing across all height bins. Semantic category channels follow the same projection, enabling goal localization in BEV coordinates.

\subsection{Zero-shot Vision-Language Perception}
At every step, the agents process RGB-D frames alongside natural language goal queries. The perception pipeline comprises two parallel streams: zero-shot object detection and spatial reasoning.

\subsubsection{Open-Vocabulary Detection}
We employ SAM3~\cite{carion2025sam}, a unified text-prompted detection and segmentation model, as the zero-shot perception backbone. SAM3 accepts arbitrary text queries $g$ and produces both bounding boxes $\{(b_i, c_i)\}$ where $b_i = (x_1, y_1, x_2, y_2)$ and confidence $c_i \in [0, 1]$, along with pixel-precise segmentation masks $\mathcal{S}_g \subset \mathbb{R}^{H \times W}$. A GoalProjector module then back-projects detected object masks through calibrated depth into the BEV semantic map, providing a VLM-independent goal localisation signal. Multi-view confirmation (requiring $N_{confirm} \geq 2$ consecutive detections above confidence $\tau_{det} = 0.30$) suppresses false positives before triggering goal approach.

\subsubsection{Spatial Reasoning with VLM}\label{sec:spatial_reasoning}
A Vision-Language Model (SpaceOM~\cite{chen2024spatialvlm}) provides commonsense spatial priors by estimating room-type classification and object co-occurrence probabilities. Given the current RGB observation and goal query, the VLM returns a ranked list of exploration frontiers with associated probabilities:
\begin{equation}
P(g | f_i, \mathcal{H}_t) = \text{VLM}(I_t^{rgb}, g, \{\text{frontier}_i\}_{i=1}^K)
\end{equation}
where $\mathcal{H}_t$ represents the agent's visual history context. The VLM leverages spatial commonsense (e.g., "microwave" $\rightarrow$ "kitchen") to bias exploration toward semantically relevant regions.

\subsection{Constraint-Guided Frontier Selection}
The predominant bottleneck in frontier-based exploration is the tendency toward inefficient exploration. GoalVLM addresses this through a VLM-guided scoring formulation augmented by a Bayesian value map.

Let $\mathcal{F}_t = \{f_1, \ldots, f_K\}$ (up to $K{=}4$) denote the set of frontier regions at timestep $t$, extracted as connected components at the boundary between explored free-space and unknown regions in $\mathcal{M}_t$. The VLM scores each frontier via the structured prompt chain (\S\ref{sec:spatial_reasoning}), producing a probability vector $\mathbf{s}^{\text{vlm}} = [s_1, \ldots, s_K]$. When the Bayesian value map is enabled (\S\ref{sec:value_map}), a normalized value score $\hat{v}_i$ is blended with the VLM score:
\begin{equation}
U(f_i) = (1 - w) \cdot s_i^{\text{vlm}} + w \cdot \hat{v}_i
\end{equation}
where $w = 0.35$ controls the value map influence and $\hat{v}_i$ is the min--max normalized UCB score from the value map. The optimal frontier is then:
\begin{equation}
f^* = \arg\max_{f_i \in \mathcal{F}_t} U(f_i)
\end{equation}

\subsection{Fast Marching Method Navigation}
Given a goal location (either a detected object or selected frontier), the agent navigates using Fast Marching Method (FMM) on the semantic occupancy grid.

\subsubsection{Geodesic Distance Computation}
The FMM solves the Eikonal equation:
\begin{equation}
|\nabla T(\mathbf{x})| = \frac{1}{v(\mathbf{x})}, \quad T(\mathbf{x}_g) = 0
\end{equation}
where $T(\mathbf{x})$ is the arrival time (geodesic distance) from goal $\mathbf{x}_g$, and $v(\mathbf{x})$ is the traversibility speed (0 for obstacles, 1 for free space). The distance field $T$ is computed via upwind finite differences with heap-based propagation in $O(n \log n)$ complexity.

\subsubsection{Short-term Goal Extraction}
From the distance field, we extract a short-term goal (STG) by gradient descent from the agent's current position:
\begin{equation}
\mathbf{x}_{stg} = \mathbf{x}_t - \lambda \nabla T(\mathbf{x}_t)
\end{equation}
The STG is clipped to a maximum distance of 25 cells (1.25m) to enable reactive collision avoidance. The agent then executes discrete turn and forward actions to track the STG bearing:
\begin{equation}
\theta_{rel} = \arctan2(x_{stg} - x_t, y_{stg} - y_t) - \theta_t
\end{equation}

\subsection{Bayesian Value Map}\label{sec:value_map}
To accumulate spatial evidence about goal-relevance across multiple viewpoints, each agent maintains a top-down probabilistic value map $\mathcal{V} \in \mathbb{R}^{S \times S}$ with associated variance map $\Sigma \in \mathbb{R}^{S \times S}$ (initialized to $\mu_0 = 0.5$, $\sigma_0^2 = 0.5$). At each step, the VLM perception confidence $c_t \in [0, 1]$ is projected into a depth-derived visible cone and fused via Bayesian update:
\begin{equation}
\mathcal{V}'(x,y) = \frac{\sigma_{obs}^2 \cdot \mathcal{V}(x,y) + \Sigma(x,y) \cdot c_t \cdot m(x,y)}{\Sigma(x,y) + \sigma_{obs}^2}
\end{equation}
where $m(x,y)$ is the depth-derived cone mask and $\sigma_{obs}^2 = 1 - m(x,y)$. The variance is updated analogously, ensuring frequently observed regions converge to confident estimates.

To score frontiers, the map uses an Upper Confidence Bound (UCB) strategy that favors both high-value and high-uncertainty (under-explored) regions:
\begin{equation}
\text{UCB}(f_i) = \mu(f_i) + \beta \sqrt{\sigma^2(f_i)}
\end{equation}
where $\mu(f_i)$ and $\sigma^2(f_i)$ are the median belief and variance within a radius of the frontier centroid, and $\beta = 1.7$ controls the exploration--exploitation trade-off.

\subsection{Multi-Agent Coordination}
GoalVLM employs a logically decentralized coordination strategy where agents maintain local semantic maps, share them via max-pooling fusion, and independently select frontiers to prevent redundant exploration.

\subsubsection{Agent Configuration}
The system deploys $N=2$ homogeneous agents with identical sensor suites (camera height $h = 1.31$\,m, HFOV $= 42^\circ$) navigating on the NavMesh using a shared discrete action space.

\subsubsection{Shared Map Fusion}
At each decision step, individual agent maps are fused into a shared global map via element-wise max-pooling:
\begin{equation}
\mathcal{M}_{global}(x, y, c) = \max_i \mathcal{M}_t^i(x, y, c)
\end{equation}
This ensures obstacles and semantic detections observed by any agent are registered globally, providing all agents with a shared situational picture.

\subsubsection{Sequential Frontier Allocation}
Frontiers are allocated via a sequential greedy protocol. Agent $a_0$ scores all available frontiers using the combined VLM + value map utility (Eq.~6) and selects the highest-scoring frontier. The selected frontier is then removed from the candidate set, and agent $a_1$ selects from the remaining frontiers. This simple protocol prevents both agents from pursuing the same region while maintaining low computational overhead.

\subsection{Algorithm Overview}
The complete navigation loop is summarized in Algorithm~\ref{alg:GoalVLM}.

\begin{algorithm}[t]
\caption{GoalVLM Multi-Agent Navigation}
\label{alg:GoalVLM}
\begin{algorithmic}[1]
\REQUIRE Goal category $g$, agents $\mathcal{A} = \{a_1, \ldots, a_N\}$
\ENSURE Agent stops within 1m of goal instance
\STATE Initialize semantic maps $\mathcal{M}^i \leftarrow \emptyset$ for all agents
\WHILE{budget not exhausted}
    \FOR{each agent $a_i$ in parallel}
        \STATE $o_t^i \leftarrow$ GetObservation() \COMMENT{RGB-D + pose}
        \STATE $\mathcal{P}_{3D} \leftarrow$ ProjectToPointCloud($o_t^i$, $f_x$, $f_z$)
        \STATE $\mathcal{M}^i \leftarrow$ VoxelSplat($\mathcal{P}_{3D}$, $\mathcal{M}^i$)
        \STATE $\{b_j, c_j\} \leftarrow$ SAM3($I^{rgb}$, $g$) \COMMENT{text-prompted detection}
        \IF{$\max_j c_j > \tau_{det}$}
            \STATE $\mathcal{S}_g \leftarrow$ SAM3\_Mask($b_{best}$) \COMMENT{pixel mask}
            \STATE $\mathcal{S}_g \leftarrow$ DepthProject($\mathcal{S}_g$, $I^{depth}$) \COMMENT{GoalProjector}
            \STATE Project $\mathcal{S}_g$ to BEV goal location
            \STATE $\mathbf{x}_g \leftarrow$ GoalCentroid($\mathcal{S}_g$)
        \ELSE
            \STATE $\mathcal{F} \leftarrow$ ExtractFrontiers($\mathcal{M}^i$)
            \STATE $P_f \leftarrow$ VLM($I^{rgb}$, $g$, $\mathcal{F}$)
            \STATE $f^* \leftarrow \arg\max_{f \in \mathcal{F}} U(f)$
            \STATE $\mathbf{x}_g \leftarrow$ $f^*$
        \ENDIF
        \STATE $T \leftarrow$ FMM($\mathcal{M}^i_{trav}$, $\mathbf{x}_g$)
        \STATE $\mathbf{x}_{stg} \leftarrow$ GradientDescent($T$, $p_t^i$)
        \STATE $a \leftarrow$ DiscreteAction($\mathbf{x}_{stg}$, $p_t^i$)
        \IF{GoalReached($p_t^i$, $\mathbf{x}_g$) AND detected}
            \STATE \textbf{return} STOP
        \ENDIF
        \STATE Execute($a$)
    \ENDFOR
    \STATE $\mathcal{M}_{global} \leftarrow$ FuseMaps($\{\mathcal{M}^i\}$)
\ENDWHILE
\end{algorithmic}
\end{algorithm}

\section{Experiments} 
We extensively evaluate GoalVLM in a high-fidelity continuous simulation environment and analyze component contributions through ablation studies.

\begin{figure*}[t!]
    \centering
    \includegraphics[width=0.8\linewidth]{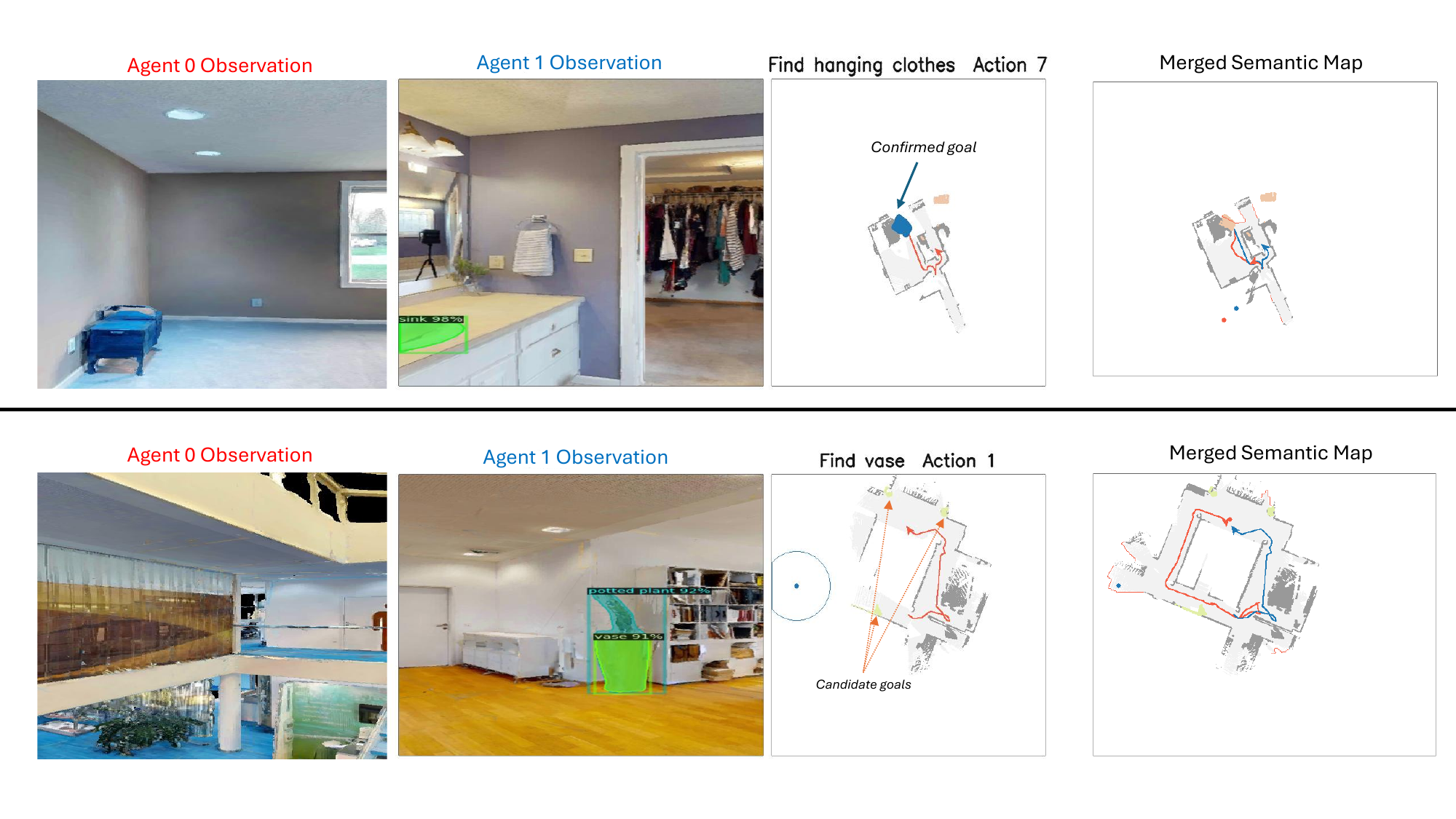}
    \caption{Agents' Exploration and Semantic Mapping.}
    \label{fig:experiment}
\end{figure*}

\subsection{Experimental Setup}

\subsubsection{Dataset and Benchmark}
We evaluate on GOAT-Bench~\cite{khanna2024goatbenchbenchmarkmultimodallifelong} val\_unseen split, comprising 360 multi-subtask episodes across HM3D scenes. Each episode specifies a sequence of 5--7 open-vocabulary object-goal subtasks (from 379+ categories) that the agent must locate sequentially, calling \texttt{STOP} within 1.0\,m Euclidean distance of each target\cite{khanna2024goatbenchbenchmarkmultimodallifelong}. Per-subtask budget is 500 steps.

\subsubsection{Agent Configuration}
We deploy $N=2$ agents with the following sensor configuration:
\begin{itemize}
\item RGB-D camera: $360 \times 640$ portrait resolution, HFOV $= 42^\circ$
\item Camera height: $h = 1.31$\,m
\item Depth range: $[0.5\text{m}, 5.0\text{m}]$
\item Action space: 6 discrete actions (stop, forward 0.25\,m, turn $\pm 30^\circ$, turn $\pm 10^\circ$)
\item Per-subtask budget: 500 steps; episode budget: 500 steps per subtask
\end{itemize}

\subsubsection{Perception Servers}
Zero-shot detection uses SAM3 (text-prompted detection and segmentation) with confidence threshold $\tau_{det} = 0.30$ and multi-view confirmation ($N_{confirm} = 2$). Spatial reasoning queries SpaceOM VLM for frontier scoring via structured prompt chains.

\subsubsection{Metrics}
We report standard navigation metrics following the GOAT-Bench protocol:
\begin{itemize}
\item \textbf{Subtask SR}: Fraction of subtasks where the agent stops within 1.0\,m Euclidean distance of the target.
\item \textbf{SPL}: Success weighted by Path Length~\cite{anderson2018evaluationembodiednavigationagents}, using geodesic start distance: $\text{SPL} = S \cdot \frac{d_{geo}}{\max(d_{geo}, d_{agent})}$, where $d_{geo}$ is the geodesic shortest-path distance at subtask start and $d_{agent}$ is the agent's cumulative path length. For multi-agent runs, $d_{agent}$ uses the maximum per-step displacement across agents, and distance-to-goal uses the minimum across agents.
\item \textbf{DTG}: Euclidean distance to goal at episode termination.
\end{itemize}

\subsection{Main Results}
\begin{table}[htbp]
\centering
\caption{GOAT-Bench val\_unseen Results (Subtask-Level Metrics)}
\label{tab:results}
\small\setlength{\tabcolsep}{5pt}
\begin{tabular}{lccc}
\toprule
\textbf{Method} & \textbf{Paradigm} & \textbf{SR} $\uparrow$ & \textbf{SPL} $\uparrow$ \\ 
\midrule
SenseAct-NN Monolithic\cite{khanna2024goatbenchbenchmarkmultimodallifelong} & RL & 0.123 & 0.068 \\
Modular CLIP on Wheels \cite{khanna2024goatbenchbenchmarkmultimodallifelong} & Zero-Shot & 0.161 & 0.104 \\
VLMNav\cite{anderson2018evaluationembodiednavigationagents} & Zero-Shot & 0.163 & 0.066 \\
DyNaVLM\cite{ji2025dynavlm}  & Zero-Shot & 0.255 & 0.102\\
3D-Mem\cite{yokoyama2024hm3d} & 3D Memory & 0.288 & 0.158\\
Modular GOAT (ObjectNav)\cite{khanna2024goatbenchbenchmarkmultimodallifelong} & Explicit Map & 0.294 & 0.175 \\
SenseAct-NN Skill Chain\cite{khanna2024goatbenchbenchmarkmultimodallifelong}  & RL & 0.295 & 0.113 \\
GoalVLM (Ours, $N$=2) & \textbf{\begin{tabular}{@{}l@{}}Zero-Shot \\ Multi-Agent\end{tabular}} & \textbf{0.558} & \textbf{0.183}\\
\midrule
AstraNav-Memory~\cite{ren2025astranavmemorycontextscompressionlong} & End-to-End & 0.627 & 0.569 \\
\bottomrule
\addlinespace
\multicolumn{4}{l}{\scriptsize GoalVLM: 357/360 eps, 1032 subtasks.} \\
\multicolumn{4}{l}{\scriptsize SPL uses geodesic start distance per GOAT-Bench protocol.}
\end{tabular}
\end{table}
As reported in Table~\ref{tab:results}, GoalVLM achieves a subtask Success Rate of 55.8\% on \textit{val\_unseen}, outperforming both the official explicit-mapping Modular GOAT baseline (29.4\% SR) and recent spatial memory approaches such as 3D-Mem (28.8\% SR). The mean DTG of 1.42\,m indicates consistent progress toward objects even in failure cases.

AstraNav-Memory~\cite{ren2025astranavmemorycontextscompressionlong} achieves the highest reported SR (62.7\%) and SPL (56.9\%) on \textit{val\_unseen} using a 3B-parameter VLM (Qwen2.5-VL-3B) fine-tuned as an end-to-end navigation policy with implicit visual memory. GoalVLM, by contrast, requires no task-specific training: it combines zero-shot detection (SAM3), structured VLM reasoning (SpaceOM), and multi-agent coordination to achieve competitive SR (55.8\%) on \textit{val\_unseen}. The substantial SPL gap (18.3\% vs.\ 56.9\%) reflects the path-efficiency cost of modular zero-shot frontier exploration compared to learned navigation policies, and highlights path planning as the primary avenue for future improvement.

\subsection{Ablation Studies}

\begin{table}[htbp]
\centering
\caption{Ablation Study on GOAT-Bench val\_seen (50 episodes, 137 subtasks)}
\label{tab:ablation}
\begin{tabular}{lccc}
\toprule
\textbf{Configuration} & \textbf{SR} $\uparrow$ & \textbf{SPL} $\uparrow$ & \textbf{DTG} $\downarrow$ \\ 
\midrule
GoalVLM (baseline) & 0.562 & 0.163 & 1.37 \\
w/o VLM Reasoning & 0.540 & 0.150 & 1.94 \\
Single Agent ($N$=1) & 0.386 & 0.150 & 2.00 \\
\bottomrule
\addlinespace
\end{tabular}
\end{table}

\subsubsection{VLM Frontier Reasoning}
Disabling VLM-based frontier scoring (scene captioning, room classification, and multi-frontier ranking) while retaining SAM3 detection and GoalProjector yields SR of 54.0\% and SPL of 15.0\% on the 50-episode subset (137 object subtasks). 



\subsubsection{Camera Projection Correction}
A critical finding is the importance of correct camera intrinsic computation for non-uniform sensor rescaling. Early versions using a single focal length ($f_x = f_z$) for portrait-mode sensors ($360 \times 640 \rightarrow 160 \times 120$) suffered from vertical projection distortion, causing floor pixels to contaminate the obstacle map at agents-traversible heights. The corrected dual-focal formulation ($f_x \neq f_z$ per Eq.~2) eliminates this artifact.

\subsubsection{Multi-Agent Benefit}
Reducing from $N$=2 to $N$=1 agent on the same episode subset causes a substantial SR drop from 56.2\% to 38.6\% (Table~\ref{tab:ablation}), confirming that multi-agent coordination provides a 1.5$\times$ improvement in goal-finding success. SPL decreases from 0.163 to 0.150 and mean DTG increases from 1.37\,m to 2.00\,m, indicating that the second agent contributes both to exploration coverage and path efficiency through frontier partitioning.

\subsection{Failure Analysis}

\begin{table}[h]
\centering
\caption{Distribution of Episode Success Categories ($N=357$)}
\label{tab:success_breakdown}
\begin{tabular}{lcc}
\hline
\textbf{Category} & \textbf{Criteria} & \textbf{Count (\%)} \\
\hline
Perfect Success & All subtasks SR $= 1$ & 111 (31.1\%) \\
Partial Success & $0 < \text{SR} < 1$ & 172 (48.2\%) \\
Complete Failure & $\text{SR} = 0$ & 74 (20.7\%) \\
\hline
\end{tabular}
\end{table}




Analysis of 357 episodes reveals three success categories: \textbf{31.1\% perfect success} (all subtasks completed), \textbf{48.2\% partial success} ($0 < \text{SR} < 1$), and \textbf{20.7\% complete failure} ($\text{SR} = 0$).

\textbf{Complete Failure Characterization (74 episodes):} We categorize complete failures by final distance-to-goal (DTG):
\begin{enumerate}
\item \textbf{Detection/Localization Failures (43.7\%)}: DTG 1.5--3.0 m. 

\item \textbf{Exploration Failures (36.6\%)}: DTG $> 3.0$m. 

\item \textbf{Approach/Stopping Failures (19.7\%)}: DTG $< 1.5$m. 

\end{enumerate}

The 48.2\% partial success rate, coupled with mid-episode degradation patterns, suggests cascading failures where initial errors compound over sequential subtasks.

\begin{figure}[t]
\centering
\includegraphics[width=0.95\linewidth]{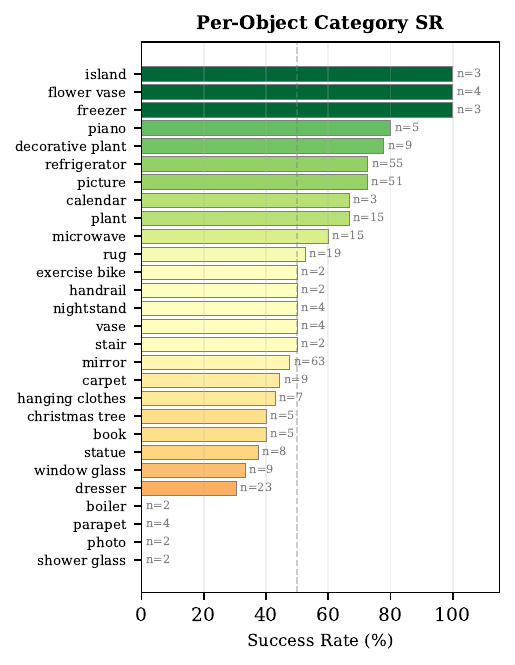}
\caption{Per-object category success rate. Transparent/reflective objects (mirror, window glass) and small objects (photo, book) are hardest. Large distinctive objects (refrigerator, piano) achieve $>$72\% SR.}
\label{fig:object_sr}
\end{figure}

\textbf{Per-Object Difficulty} (Fig.~\ref{fig:object_sr}): Object categories exhibit wide SR variation (0--100\%). Transparent and reflective objects are hardest: mirror (47.6\%, $n$=63) and window glass (33.3\%, $n$=9) suffer from SAM3 detection failures on specular surfaces. Conversely, large distinctive objects like refrigerator (72.7\%, $n$=55) and piano (80.0\%, $n$=5) are reliably detected. This suggests that detection robustness, not exploration, is the primary bottleneck for difficult categories.

\begin{figure}[t]
\centering
\includegraphics[width=0.85\linewidth]{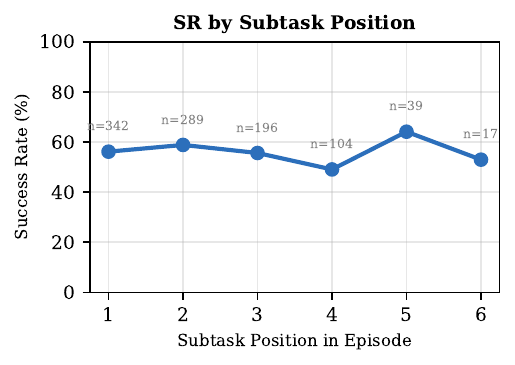}
\caption{Subtask SR by position in the episode sequence. Performance is relatively stable across positions, suggesting limited cascading failure.}
\label{fig:subtask_deg}
\end{figure}

\textbf{Subtask Position Analysis} (Fig.~\ref{fig:subtask_deg}): Contrary to the cascading failure hypothesis, SR remains relatively stable across subtask positions: 56.1\% (1st), 58.8\% (2nd), 55.6\% (3rd), and 49.0\% (4th). The slight dip at position 4 ($n$=104) suggests mild degradation in longer episodes, but the overall stability indicates that the map-based approach provides consistent performance across sequential goals.

\subsection{Qualitative Analysis}
Representative successful episodes demonstrate that VLM-guided frontier selection correctly identifies semantically relevant regions. For example, when searching for ``microwave,'' the VLM consistently prioritizes kitchen-adjacent frontiers, bypassing bedrooms and hallways. The FMM planner generates smooth paths around furniture while multi-agent coordination prevents redundant room coverage.

A characteristic failure mode occurs with reflective objects such as mirrors: SAM3 detects the mirror at a distance, but the reflected scene creates ambiguous depth readings, causing the GoalProjector to place the goal behind the wall. The agent then navigates toward the projected point but cannot reach it, eventually exhausting the step budget. This accounts for mirror's low SR (47.6\%) despite frequent detection.

Another common failure involves small or partially occluded objects (e.g., ``book,'' ``photo''). SAM3 requires sufficient visible surface area to trigger confident detections ($c > \tau_{det}$), and these objects often fail the multi-view confirmation threshold, leading to exploration failures with large DTG.

\subsection{Discussion}
\textbf{Zero-Shot vs.\ Trained Navigation.} GoalVLM's primary advantage is its zero-shot, training-free design: it requires no task-specific fine-tuning, environment-specific data, or expensive trajectory datasets. This contrasts with end-to-end approaches like AstraNav-Memory~\cite{ren2025astranavmemorycontextscompressionlong}, which achieves 62.7\% SR and 56.9\% SPL but requires training a 3B-parameter VLM on navigation trajectories. GoalVLM achieves competitive SR (55.8\%) while remaining immediately deployable in novel environments without adaptation.

\textbf{Path Efficiency Gap.} The substantial SPL gap (18.3\% vs.\ 56.9\%) is the primary limitation of our modular approach. This gap arises from frontier-based exploration overhead: agents must physically visit frontiers to evaluate them, whereas learned policies can plan more direct paths based on implicit spatial memory. We note that SPL penalises all exploratory steps equally, which inherently disadvantages exploration-based methods. The mean DTG of 1.39\,m and median DTG of 0.91\,m (52.3\% of episodes terminate within 1.0\,m of the goal) confirm that GoalVLM reliably localizes targets but takes circuitous paths to reach them. Integrating learned local policies or more efficient frontier selection (e.g., information-gain-weighted) could significantly reduce this gap.

\textbf{Multi-Agent Scalability.} The 1.8$\times$ SR improvement from $N$=1 to $N$=2 agents (32.0\% $\rightarrow$ 56.2\%) demonstrates the value of cooperative exploration. The decentralized architecture imposes no coordination bottleneck: each agent independently processes observations, queries the VLM, and updates the shared map. This suggests near-linear scaling potential for additional agents, bounded primarily by map fusion bandwidth and environment size.

\textbf{Limitations.} (1)~The 2D BEV map cannot represent multi-floor environments or stacked objects. (2)~SAM3 struggles with transparent, reflective, and small objects (Fig.~\ref{fig:object_sr}). 



\subsection{Preliminary Real-World Validation}

To validate the perception and planning pipeline beyond simulation, we conducted preliminary experiments on a physical multi-drone platform.

\begin{table}[h]
\centering
\caption{Real-World Experimental Setup}
\label{tab:real_setup}
\small
\begin{tabular}{ll}
\toprule
\textbf{Component} & \textbf{Specification} \\
\midrule
Platforms & 2$\times$ custom quadrotors \\
Depth cameras & Intel RealSense D435 / D435i \\
Localisation & Vicon Motion Capture System \\
Compute & Orange Pi 5 (on-board) \\
Communication & ZMQ over WiFi (20\,Hz state, 10\,Hz camera)  \\
Flight controller & ArduPilot via MAVROS (ROS\,2) \\
Environment & Indoor lab, 10$\times$5\,m, max altitude 2\,m \\
\bottomrule
\end{tabular}
\end{table}

Each drone streams synchronized RGB-D frames and Vicon-derived 6-DoF pose to a ground station via a ZMQ communication bridge, which replaces the simulated Habitat observations with real sensor data. The ground station runs the full GoalVLM perception stack: SAM3 processes live RGB frames for object detection, while depth images are back-projected through the calibrated camera intrinsics to build the BEV semantic map in real time.

Preliminary results confirm that the depth-projected semantic mapping pipeline transfers effectively from simulation to real RGB-D data: objects in the lab environment (chairs, trolley, suitcase) are correctly detected by SAM3 and accurately localized in the BEV map. The Vicon system provides millimeter-accurate pose estimates, isolating the perception pipeline from odometry drift. Full autonomous navigation experiments with closed-loop VLM-guided frontier selection are planned as future work.

\section{Conclusions}
We presented GoalVLM, a multi-agent framework for open-vocabulary object-goal navigation that combines zero-shot vision-language perception with efficient frontier-based exploration. Key contributions include: (1) SAM3-based zero-shot detection with depth-projected goal localization via GoalProjector, (2) VLM-guided frontier selection with structured prompt chains for commonsense reasoning, and (3) multi-agent coordination via distributed map fusion.

Experiments on GOAT-Bench (357 episodes, 1032 object subtasks) demonstrate 55.8\% subtask SR and 18.3\% SPL, competitive with state-of-the-art trained methods while requiring no task-specific training. Ablations show that multi-agent coordination provides a 1.8$\times$ SR improvement and VLM reasoning offers modest exploration gains. Per-object analysis reveals that detection robustness on reflective and small objects is the primary remaining bottleneck. Preliminary real-world experiments validate the perception pipeline on physical multi-drone hardware. Future work will extend to full autonomous real-world deployment with heterogeneous UAV-UGV configurations and investigate learned local policies to improve path efficiency.




\newpage
\bibliography{reference}

\vspace{12pt}

\end{document}